\documentclass[11pt]{article}
\usepackage{acl2016}
\usepackage{times}
\usepackage{latexsym}
\usepackage{multirow, graphicx, color, url}
\usepackage{subcaption}

\aclfinalcopy 

\newcommand{\NOTE}[1]{} 
\newcommand{\KA}[1]{}

\title{Integrating Distributional Lexical Contrast into Word Embeddings for Antonym--Synonym Distinction}

\author{Kim Anh Nguyen \and Sabine Schulte im Walde \and Ngoc Thang Vu \\
	    Institut f\"ur Maschinelle Sprachverarbeitung\\
	    Universit\"at Stuttgart\\
	    Pfaffenwaldring 5B, 70569 Stuttgart, Germany\\
	    {\{\tt nguyenkh,schulte,thangvu\}@ims.uni-stuttgart.de}}

\date{}

\begin{document}

\maketitle

\begin{abstract}
  We propose a novel vector representation that integrates lexical
  contrast into distributional vectors and strengthens the most
  salient features for determining degrees of word similarity. The
  improved vectors significantly outperform standard models and
  distinguish antonyms from synonyms with an average precision of
  0.66--0.76 across word classes (adjectives, nouns, verbs). Moreover,
  we integrate the lexical contrast vectors into the objective
  function of a skip-gram model. The novel embedding outperforms
  state-of-the-art models on predicting word similarities in
  SimLex-999, and on distinguishing antonyms from synonyms.
\end{abstract}

\section{Introduction}

Antonymy and synonymy represent lexical semantic relations that are
central to the organization of the mental
lexicon~\cite{Miller/Fellbaum:91}. While antonymy is defined as the
oppositeness between words, synonymy refers to words that are similar
in meaning~\cite{Deese:65,Lyons1977}. From a computational point of
view, distinguishing between antonymy and synonymy is important for
NLP applications such as Machine Translation and Textual Entailment,
which go beyond a general notion of semantic relatedness and require
to identify specific semantic relations. However, due to
interchangeable substitution, antonyms and synonyms often occur in
similar contexts, which makes it challenging to automatically
distinguish between them.

Distributional semantic models (DSMs) offer a means to represent
meaning vectors of words and to determine their semantic
``relatedness'' \cite{Budanitsky/Hirst:06,Turney/Pantel2010}. They
rely on the \textit{distributional
  hypothesis}~\cite{Harris1954,Firth:57}, in which words with similar
distributions have related meaning. For computation, each word is
represented by a weighted feature vector, where features typically
correspond to words that co-occur in a particular context.
However, DSMs tend to retrieve both synonyms (such as
\textit{formal--conventional}) and antonyms (such as
\textit{formal--informal}) as related words and cannot sufficiently
distinguish between the two relations. 

In recent years, a number of distributional approaches have accepted
the challenge to distinguish antonyms from synonyms, often in
combination with lexical resources such as thesauruses or
taxonomies. For example, \newcite{Lin2003} used dependency triples to
extract distributionally similar words, and then in a post-processing
step filtered out words that appeared with the patterns `from X to Y'
or `either X or Y' significantly often. \newcite{Mohammad2013} assumed
that word pairs that occur in the same thesaurus category are close in
meaning and marked as synonyms, while word pairs occurring in
contrasting thesaurus categories or paragraphs are marked as
opposites. \newcite{Scheible2013} showed that the distributional
difference between antonyms and synonyms can be identified via a
simple word space model by using appropriate features.
\newcite{Santus2014b} and \newcite{Santus2014} aimed to identify the
most salient dimensions of meaning in vector representations and
reported a new average-precision-based distributional measure and an
entropy-based measure to discriminate antonyms from synonyms (and
further paradigmatic semantic relations).

Lately, antonym--synonym distinction has also been a focus of word
embedding models. For example, \newcite{Adel2014} integrated
coreference chains extracted from large corpora into a skip-gram model
to create word embeddings that identified antonyms. \newcite{Ono2015}
proposed thesaurus-based word embeddings to capture antonyms. They
proposed two models: the WE-T model that trains word embeddings on
thesaurus information; and the WE-TD model that incorporated
distributional information into the WE-T model. \newcite{Nghia2015}
introduced the multitask lexical contrast model (mLCM) by
incorporating WordNet into a skip-gram model to optimize semantic
vectors to predict contexts. Their model outperformed standard
skip-gram models with negative sampling on both general semantic tasks
and distinguishing antonyms from synonyms.

In this paper, we propose two approaches that make use of lexical
contrast information in distributional semantic space and word
embeddings for antonym--synonym distinction. Firstly, we incorporate
lexical contrast into distributional vectors and strengthen those word
features that are most salient for determining word similarities,
assuming that feature overlap in synonyms is stronger than feature
overlap in antonyms.
Secondly, we propose a novel extension of a skip-gram model with
negative sampling~\cite{Mikolov2013b} that integrates the lexical
contrast information into the objective function. The proposed model
optimizes the semantic vectors to predict degrees of word similarity
and also to distinguish antonyms from synonyms.  The improved word
embeddings outperform state-of-the-art models on antonym--synonym
distinction and a word similarity task.

\section{Our Approach}

In this section, we present the two contributions of this paper: a new
vector representation that improves the quality of weighted features
to distinguish between antonyms and synonyms
(Section~\ref{sec:weights}), and a novel extension of skip-gram models
that integrates the improved vector representations into the objective
function, in order to predict similarities between words and to
identify antonyms (Section~\ref{sec:skip}).

\subsection{Improving the weights of feature vectors}
\label{sec:weights}

We aim to improve the quality of weighted feature vectors by
strengthening those features that are most salient in the vectors and
by putting less emphasis on those that are of minor importance, when
distinguishing degrees of similarity between words. We start out with
standard corpus co-occurrence frequencies and apply \textit{local
  mutual information (LMI)} \cite{Evert2005} to determine the original
strengths of the word features. Our score $weight^{SA}(w,f)$
subsequently defines the weights of a target word $w$ and a feature
$f$:
\vspace{+1mm}\\
\begin{equation}
  \label{eq:weight-SA}
  \resizebox{0.99\hsize}{!}{%
  $\begin{array}{r}
    weight^{SA}(w,f) = \frac{1}{{\# (w,u)}}\sum\nolimits_{u \in W(f) \cap S(w)} {sim(w,u)} \\
    - \frac{1}{{\# (w',v)}}\sum\limits_{w' \in A(w)}{\sum\nolimits_{v \in W(f) \cap S(w')} {sim(w',v)}}
  \end{array}$
  }
\end{equation}
\vspace{+1mm}\\
The new $weight^{SA}$ scores of a target word $w$ and a feature $f$
exploit the differences between the average similarities of synonyms
to the target word ($sim(w,u)$, with $u \in S(w)$), and the average
similarities between antonyms of the target word ($sim(w',v)$, with
$w' \in A(w)$ and $v \in S(w')$). Only those words $u$ and $v$ are
included in the calculation that have a positive original LMI score
for the feature $f$: $W(f)$. To calculate the similarity $sim$ between
two word vectors, we rely on cosine distances.
If a word $w$ is not associated with any synonyms or antonyms in our
resources (cf. Section~\ref{sec:data}), or if a feature does not
co-occur with a word $w$, we define $weight^{SA}(w,f) = 0$.

\begin{figure*}[t]
\includegraphics[width=\textwidth]{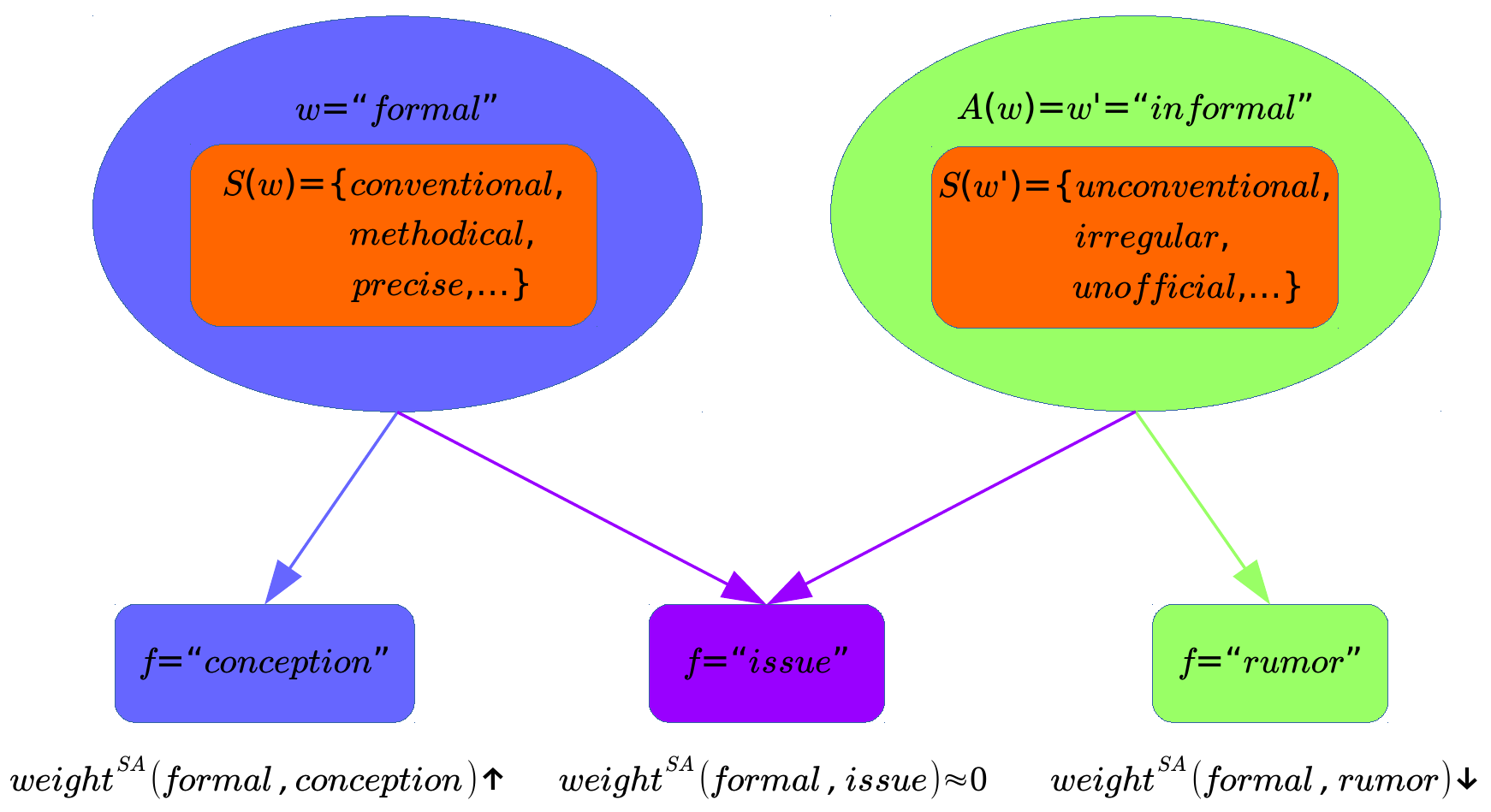}
\caption{Illustration of the $weight^{SA}$ scores for the adjective
  target \textit{formal}. The feature \textit{conception} only occurs
  with \textit{formal} and synonyms of \textit{formal}, so
  $weight^{SA}(formal, conception)$ should return a positive value;
  the feature \textit{rumor} only occurs with the antonym
  \textit{informal} and with synonyms of \textit{informal}, so
  $weight^{SA}(formal, rumor)$ should return a negative value; the
  feature \textit{issue} occurs with both \textit{formal} and
  \textit{informal} and also with synonyms of these two adjectives, so
  $weight^{SA}(formal, issue)$ should return a feature score near
  zero.}
\label{weight-graph}
\end{figure*}

The intuition behind the \textit{lexical contrast information} in our new $weight^{SA}$ is as
follows. The strongest features of a word also tend to represent
strong features of its synonyms, but weaker features of its
antonyms. For example, the feature \textit{conception} only occurs
with synonyms of the adjective \textit{formal} but not with the
antonym \textit{informal}, or with synonyms of the antonym
\textit{informal}.
$weight^{SA}(formal,conception)$, which is calculated as the average
similarity between \textit{formal} and its synonyms minus the average
similarity between \textit{informal} and its synonyms, should thus
return a high positive value. In contrast, a feature such as
\textit{issue} that occurs with many different adjectives, would
enforce a feature score near zero for $weight^{SA}(formal,issue)$,
because the similarity scores between \textit{formal} and its synonyms
and \textit{informal} and its synonyms should not differ
strongly. Last but not least, a feature such as \textit{rumor} that
only occurs with \textit{informal} and its synonyms, but not with the
original target adjective \textit{formal} and its synonyms, should
invoke a very low value for $weight^{SA}(formal,rumor)$. Figure
\ref{weight-graph} provides a schematic visualization for computing
the new $weight^{SA}$ scores for the target \textit{formal}.


Since the number of antonyms is usually much smaller than the number
of synonyms, we enrich the number of antonyms: Instead of using the
direct antonym links, we consider all synonyms of an antonym
$w' \in A(w)$ as antonyms of $w$. For example, the target word
\textit{good} has only two antonyms in WordNet (\textit{bad} and
\textit{evil}), in comparison to 31 synonyms. Thus, we also exploit
the synonyms of \textit{bad} and \textit{evil} as antonyms for
\textit{good}.

\subsection{Integrating the distributional lexical contrast into a skip-gram model}
\label{sec:skip}

Our model relies on Levy and Goldberg~\shortcite{Levy2014b}
who showed that the objective function for a skip-gram model with
negative sampling (SGNS) can be defined as follows:
\vspace{+1mm}\\
\begin{equation}
  \label{eq:sgns}
  \begin{array}{l}
    \sum\limits_{w \in V} {\sum\limits_{c \in V} \{  } \# (w,c)\log \sigma (sim(w,c))\\ 
    + k\# (w){P_0}(c)\log \sigma ( - sim(w,c))\}
  \end{array}
\end{equation}
\vspace{+1mm}\\
The first term in Equation~(\ref{eq:sgns}) represents the
co-occurrence between a target word $w$ and a context $c$ within a
context window. The number of appearances of the target word and that
context is defined as $\#(w,c)$. The second term refers to the
negative sampling where $k$ is the number of negatively sampled words,
and $\#(w)$ is the number of appearances of $w$ as a target word in
the unigram distribution $P_0$ of its negative context $c$.

To incorporate our lexical contrast information into the SGNS model,
we propose the objective function in Equation~(\ref{eq:dlce}) to add
distributional contrast followed by all contexts of the target
word. $V$ is the vocabulary; $\sigma (x) = \frac{1}{{1 + {e^{ - x}}}}$
is the sigmoid function; and $sim(w_1,w_2)$ is the cosine similarity
between the two embedded vectors of the corresponding two words $w_1$
and $w_2$. We refer to our distributional lexical-contrast embedding
model as \textit{dLCE}.
\begin{equation}
  \label{eq:dlce}
  \begin{array}{l}
    \sum\limits_{w \in V} {\sum\limits_{c \in V} \{  } (\# (w,c)\log \sigma (sim(w,c)) \\
    + k\# (w){P_0}(c)\log 	\sigma ( - sim(w,c)))\\
    + (\frac{1}{{\# (w,u)}}\sum\nolimits_{u \in W(c) \cap S(w)} {sim(w,u)}\\ 
    - \frac{1}{{\# (w,v)}}\sum\nolimits_{v \in W(c) \cap A(w)} {sim(w,v)} )\} 
  \end{array}
\end{equation}

Equation~(\ref{eq:dlce}) integrates the lexical contrast information
in a slightly different way compared to Equation~(\ref{eq:weight-SA}):
For each of the target words $w$, we only rely on its antonyms $A(w)$
instead of using the synonyms of its antonyms $S(w')$. This makes the
word embeddings training more efficient in running time, especially
since we are using a large amount of training data.

The dLCE model is similar to the WE-TD model~\cite{Ono2015} and the
mLCM model~\cite{Nghia2015}; however, while the WE-TD and mLCM models
only apply the lexical contrast information from WordNet to each of
the target words, dLCE applies lexical contrast to every single
context of a target word in order to better capture and classify
semantic contrast. 

\begin{table*}[t]
  \centering
    \begin{tabular}{l|ll|ll|ll}
     \hline
      \multirow{2}{*}{} & \multicolumn{2}{c|}{\textbf{Adjectives}} & \multicolumn{2}{c|}{\textbf{Nouns}} & \multicolumn{2}{c}{\textbf{Verbs}} \\ \cline{2-7} 
                        & ANT             & SYN             & ANT              & SYN             & ANT              & SYN             \\ \hline
      LMI               & 0.46			   & 0.56			 & 0.42			   & 0.60			 & 0.42			   & 0.62 \\
      $weight^{SA}$         & \textbf{0.36}$^{**}$           & \textbf{0.75}$^{**}$            & \textbf{0.40}             & \textbf{0.66}            & \textbf{0.38}$^{*}$             & \textbf{0.71}$^{*}$           \\ \hline

      LMI + SVD		& 0.46			 & 0.55			  & 0.46				& 0.55				  & 0.44			& 0.58 		\\                 
      $weight^{SA}$ + SVD  & \textbf{0.36}$^{***}$            & \textbf{0.76}$^{***}$            & \textbf{0.40}$^{*}$             & \textbf{0.66}$^{*}$            & \textbf{0.38}$^{***}$             & \textbf{0.70}$^{***}$           

    \end{tabular}
  \caption{AP evaluation on DSMs.}
  \vspace{+5mm}
  \label{DSMs-AP}
\end{table*}

\vspace{+3mm}
\section{Experiments}

\vspace{+2mm}
\subsection{Experimental Settings}
\label{sec:data}

The corpus resource for our vector representations is one of the
currently largest web corpora:
\textit{ENCOW14A}~\cite{Schaefer2012,Schaefer2015},
containing approximately 14.5 billion tokens
and 561K distinct word types. As distributional information, we used a
window size of 5 tokens for both the original vector representation
and the word embeddings models. For word embeddings models, we trained
word vectors with 500 dimensions; $k$ negative sampling was set to 15;
the threshold for sub-sampling was set to $10^{-5}$; and we ignored
all words that occurred $<100$ times in the corpus. The parameters of
the models were estimated by backpropagation of error via stochastic
gradient descent. The learning rate strategy was similar to Mikolov et
al.~\shortcite{Mikolov2013a} in which the initial learning rate was
set to 0.025. For the lexical contrast information, we used
WordNet~\cite{Miller1995} and
Wordnik\footnote{\url{http://www.wordnik.com}} to collect antonyms and
synonyms, obtaining a total of 363,309 synonym and 38,423 antonym
pairs.

\vspace{+2mm}
\subsection{Distinguishing antonyms from synonyms}
\label{sec:task1}

The first experiment evaluates our lexical contrast vectors by
applying the vector representations with the improved $weight^{SA}$
scores to the task of distinguishing antonyms from synonyms. As gold
standard resource, we used the English dataset described in
\cite{roth2014}, containing 600 adjective pairs (300 antonymous pairs
and 300 synonymous pairs), 700 noun pairs (350 antonymous pairs and
350 synonymous pairs) and 800 verb pairs (400 antonymous pairs and 400
synonymous pairs). For evaluation, we applied Average Precision
(AP)~\cite{Voorhees1999}, a common metric in information retrieval
previously used by \newcite{Kotlerman2010} and \newcite{Santus2014b},
among others.

Table~\ref{DSMs-AP}
presents the results of the first
experiment, comparing our improved vector representations with the
original LMI representations across word classes, without/with
applying singular-value decomposition (SVD), respectively. In order to
evaluate the distribution of word pairs with AP, we sorted the
synonymous and antonymous pairs by their cosine scores. A synonymous
pair was considered correct if it belonged to the first half; and an
antonymous pairs was considered correct if it was in the second
half. The optimal results would thus achieve an AP score of 1 for
$SYN$ and 0 for $ANT$. The results in the tables demonstrate that
$weight^{SA}$
significantly\footnote{$\chi^2, ^{***} p<.001, ^{**} p<.005$,
  $^* p<.05$} outperforms the original vector representations across
word classes.

\begin{figure*}
  \begin{subfigure}{.33\textwidth}
    \includegraphics[width=\textwidth]{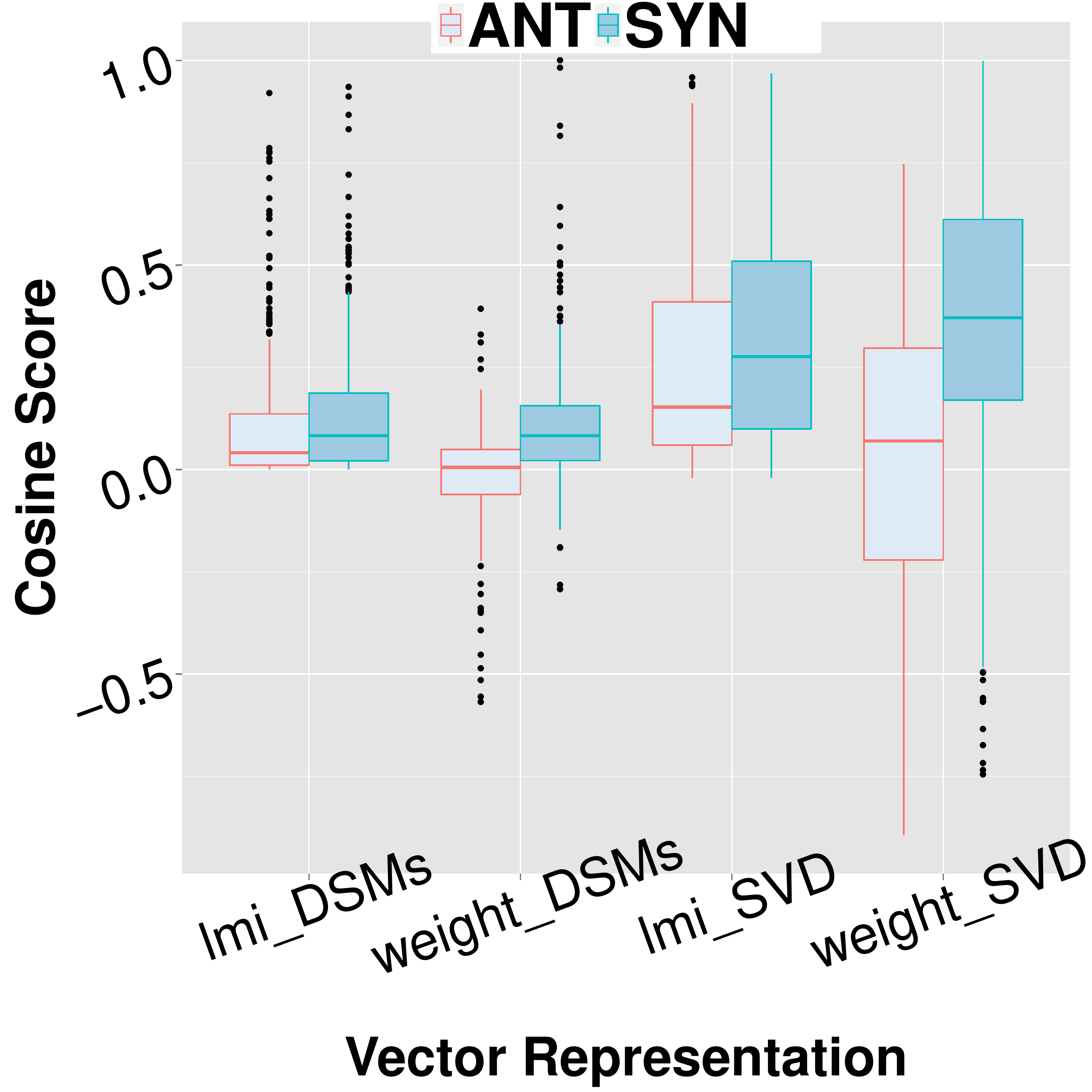}
    \caption{Cosine scores between adjectives.}
    \label{fig_adj}
  \end{subfigure}
  \begin{subfigure}{.33\textwidth}
    \includegraphics[width=\textwidth]{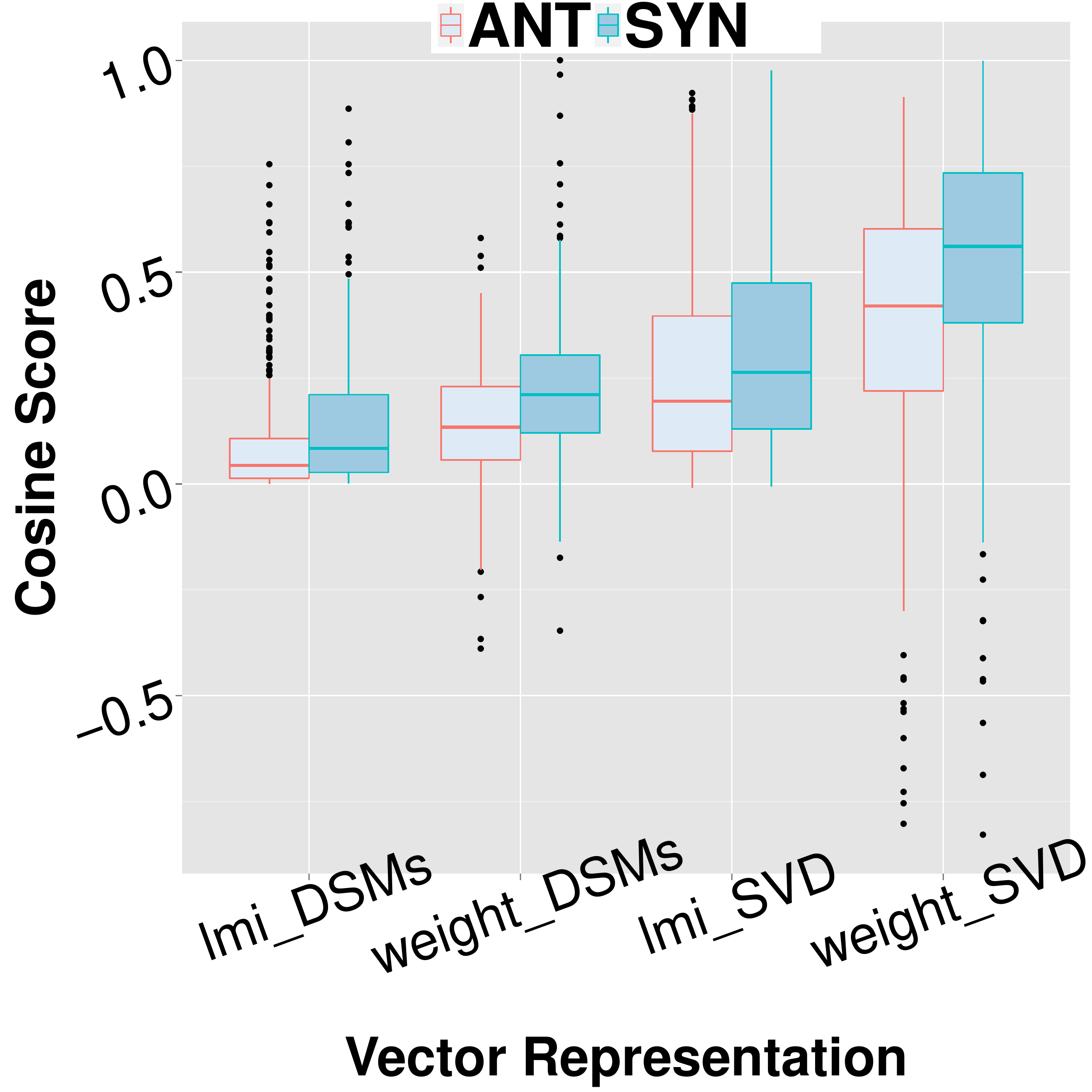}
    \caption{Cosine scores between nouns.}
    \label{fig_noun}
  \end{subfigure}
  %
  \begin{subfigure}{.33\textwidth}
    \includegraphics[width=\textwidth]{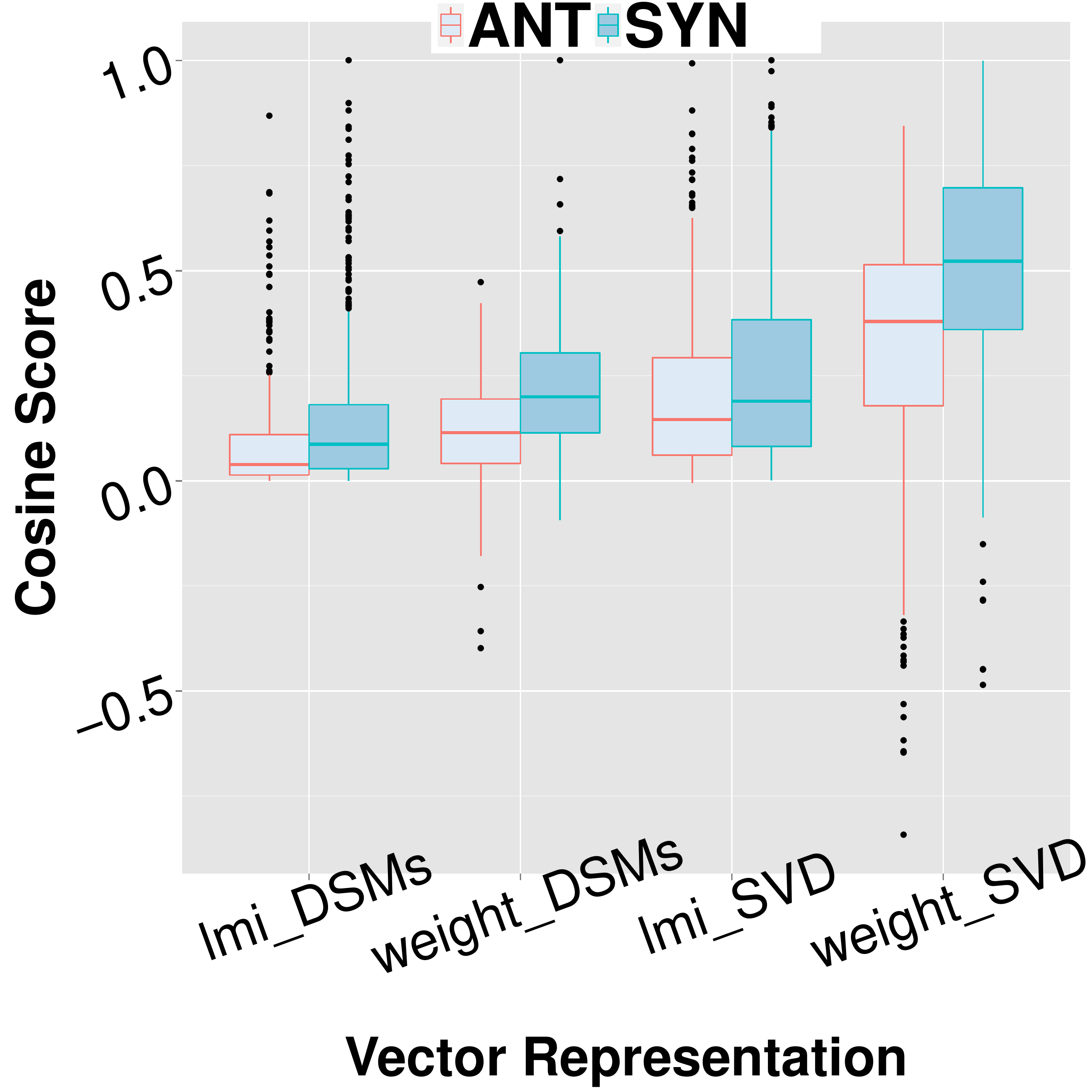}
    \caption{Cosine scores between verbs.}
    \label{fig_verb}
  \end{subfigure}
  \label{fig_boxes}
  \caption{Differences between cosine scores for antonymous vs. synonymous word pairs.}
  \vspace{+5mm}
\end{figure*}
In addition, Figure 2 compares the medians of cosine similarities
between antonymous pairs (red) vs. synonymous pairs (green) across
word classes, and for the four conditions (1) LMI, (2) $weight^{SA}$,
(3) SVD on LMI, and (4) SVD on $weight^{SA}$. The plots show that the
cosine similarities of the two relations differ more strongly with our
improved vector representations in comparison to the original LMI
representations, and even more so after applying SVD.

\vspace{+2mm}
\subsection{Effects of distributional lexical contrast on word embeddings}
\label{sec:task2}

The second experiment evaluates the performance of our dLCE model on
both antonym--synonym distinction and a word similarity task.
%
The similarity task requires to predict the degree of similarity for
word pairs, and the ranked list of predictions is evaluated against a
gold standard of human ratings, relying on the Spearman rank-order
correlation coefficient $\rho$ \cite{Siegel/Castellan:88}. 

In this paper, we use the \textit{SimLex-999} dataset~\cite{Hill2015}
to evaluate word embedding models on predicting similarities. The
resource contains 999 word pairs (666 noun, 222 verb and 111 adjective
pairs) and was explicitly built to test models on capturing similarity
rather than relatedness or association. Table~\ref{simlex999} shows
that our dLCE model outperforms both SGNS and mLCM, proving that the
lexical contrast information
has a positive effect on predicting similarity.

Therefore, the improved distinction between synonyms (strongly similar
words) and antonyms (often strongly related but highly dissimilar
words) in the dLCE model also supports the distinction between degrees
of similarity. \NOTE{Significance?}
\begin{table}[]
  \centering
  \begin{tabular}{ccc}
    \hline
    SGNS & mLCM & dLCE \\ \hline
    0.38 & 0.51 & \textbf{0.59} \\ \hline
  \end{tabular}
  \caption{Spearman's $\rho$ on SimLex-999.}
  \label{simlex999}
  \vspace{+2mm}
\end{table}

\NOTE{SIW: I don't understand this task. Where does the data come
  from?} \KA{the dataset is described in Section~\ref{sec:task1}} For
distinguishing between antonyms and synonyms, we computed the cosine
similarities between word pairs on the dataset described in
Section~\ref{sec:task1}, and then used the area under the ROC curve
(AUC) to evaluate the performance of dLCE compared to SGNS and
mLCM. The results in Table~\ref{classify_task} report that dLCE
outperforms SGNS and mLCM also on this task. \NOTE{Significance?}
\begin{table}[]
  \centering
    \begin{tabular}{l|ccc}
      \hline
                  & Adjectives & Nouns & Verbs \\ \hline
      SGNS        & 0.64          & 0.66           & 0.65           \\
      mLCM        & 0.85          & 0.69           & 0.71           \\
      dLCE       & \textbf{0.90} & \textbf{0.72}  & \textbf{0.81}          
    \end{tabular}
  \caption{AUC scores for identifying antonyms.}
  \label{classify_task}
\end{table}

\section{Conclusion}

This paper proposed a novel vector representation which enhanced the
prediction of word similarity, both for a traditional distributional
semantics model and word embeddings. Firstly, we significantly
improved the quality of weighted features to distinguish antonyms from
synonyms by using lexical contrast information. Secondly, we
incorporated the lexical contrast information into a skip-gram model
to successfully predict degrees of similarity and also to identify antonyms.

\bibliography{acl2016}
\bibliographystyle{acl2016}

\end{document}